**Where to Build Food Banks and Pantries: A Two-Level Machine Learning Approach**

Gavin Ruan,[1,2] Ziqi Guo,[3,+] and Guang Lin[3,4*]

[1]West Lafayette Jr./Sr. High School, West Lafayette, IN 47906
[2]Purdue University, West Lafayette, IN 47906
[3]School of Mechanical Engineering, Purdue University, West Lafayette, IN 47906
[4]Department of Mathematics, Purdue University, West Lafayette, IN 47906

Corresponding authors: Z.G. ([gziqi@purdue.edu](gziqi@purdue.edu)), G.L. ([guanglin@purdue.edu](guanglin@purdue.edu))

**Abstract**

Over 44 million Americans currently suffer from food insecurity, of whom 13 million are children. Across the United States, thousands of food banks and pantries serve as vital sources of food and other forms of aid for food insecure families. By optimizing food bank and pantry locations, food would become more accessible to families who desperately require it. In this work, we introduce a novel two-level optimization framework, which utilizes the K-Medoids clustering algorithm in conjunction with the Open-Source Routing Machine engine, to optimize food bank and pantry locations based on real road distances to houses and house blocks. Our proposed framework also has the adaptability to factor in considerations such as median household income using a pseudo-weighted K-Medoids algorithm. Testing conducted with California and Indiana household data, as well as comparisons with real food bank and pantry locations showed that interestingly, our proposed framework yields food pantry locations superior to those of real existing ones and saves significant distance for households, while there is a marginal penalty on the first level food bank to food pantry distance. Overall, we believe that the second-level benefits of this framework far outweigh any drawbacks and yield a net benefit result.

**1 INTRODUCTION**

Food insecurity continues to remain a longstanding challenge in the United States, with over 44 million food-insecure people in 2023, 13 million of which were children ("USDA ERS - Food Security in the U.S.," n.d.). Hunger has been linked to health issues including diabetes, high blood pressure, and heart disease (Carlson 1916). For children, food insecurity is linked with increased cases of asthma, amenia, anxiety, and aggression ("Hunger in America | Feeding America," n.d.). In addition to health problems, food insecurity is especially detrimental to those seeking medical care, housing, and education (Rich, Holroyd, and Evans 2004).

The bulk of efforts to quell food insecurity lie in organizations serving communities through food banks and food pantries. According to Feeding America, over 49 million people turned to its services for help in 2022 ("Hunger in America | Feeding America," n.d.). Despite the importance of food banks and pantries, lack of transportation or available service locations means that these resources are still not easily accessible by a substantial portion of the food insecure population (Baek 2016). Therefore, it is important to optimize the food band and pantry locations such that the total or average distance from households to their nearest food pantries is the shortest. This problem can be solved with applications of the classic facility location problem.

Prior research has shown that a facility location problem can be turned into a clustering problem, with K-Means achieving a heuristic and satisfactory solution (Liao and Guo 2008). However, the K-Means clustering algorithm exhibits a major drawback in its practical application; by only using Euclidean distance as its distance factor, K-Means does not incorporate important considerations such as driving or walking distance, or terrain. Road distance is especially important when considering that 91.7% of American households have at least one vehicle (Tilford 2023). For the facility location problem, it also chooses a final location that may not have any supporting or existing infrastructure around it. As such, an adaptable framework able to utilize real driving distance would be superior in practical applications and future implementation.

With the paramount problem of food insecurity in mind, this study seeks to demonstrate a machine learning algorithm based on a two-level K-Medoids clustering method in conjunction with a routing machine to quickly yield optimal and accessible food bank and pantry locations based on driving distance. We will describe the two datasets used and their significant differences, then explain how K-Medoids will achieve desired results on both. We then compare the food bank and pantry locations generated by our two-level K-Medoids approach against real food bank and pantry locations affiliated with Feeding America and evaluate the savings in total driving distances and the average driving distance per household. Finally, we suggest future improvements and further work that can be done.

## 2 MATERIALS AND METHODS

### 2.1 Datasets and Open-Source Routing Machine

Two main datasets were used in the development and testing of the program. The first dataset used consisted of Indiana residences from a 2020 GIS survey, taken from the Indiana Geographical Information System website, a state government-affiliated database. This dataset consisted of any place of residency, including houses and apartments. As this dataset does not include any relevant data aside from longitude and latitude coordinates, a standard non-weighted K-Medoids algorithm was used. The initial dataset consisted of 3,146,263 places of residence as seen in Figure 1 (a). The second was a set of house blocks in California from the 1990 census, consisting of longitude and latitude coordinates, and median income of surrounding houses. Therefore, we included the median income as an additional parameter for running a weighted K-Medoids algorithm, which will be discussed in the K-Medoids Implementation section. All households with annual income above $40,000 were removed from consideration, to prevent skewed placement by households with likely little need for food aid. After all modifications were applied, 12,533 households remained in the dataset as seen from Figure 1 (b).

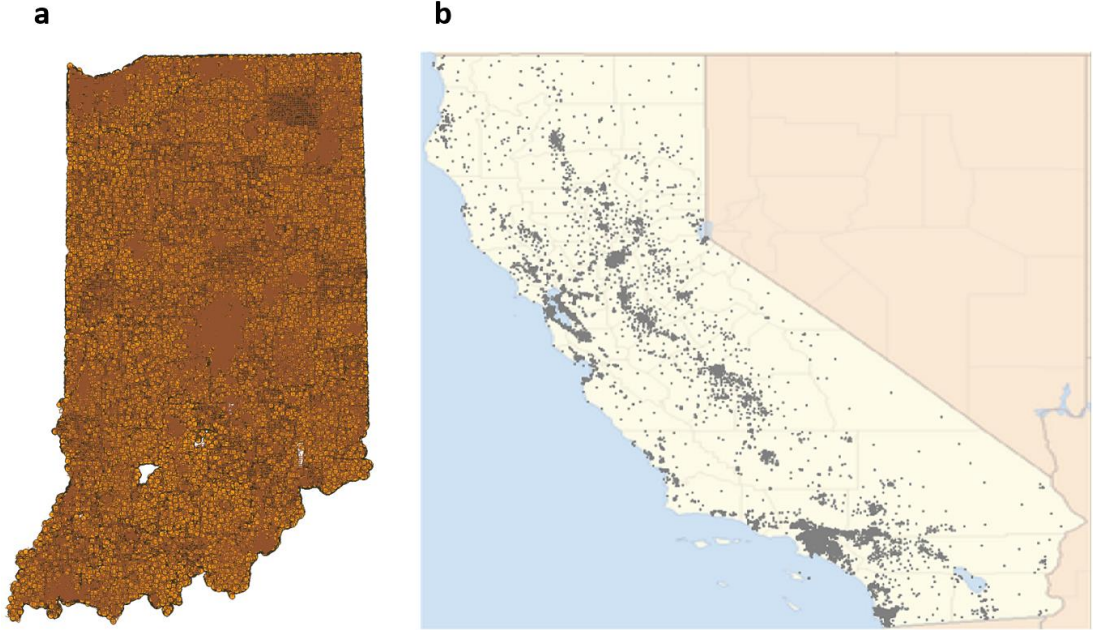

**Figure 1** (a) The considered Indiana residences from a 2020 GIS survey. (b) The considered California households from a 1990 census.

To consider real driving distance, we utilized the Open-Source Routing Machine (OSRM), an application which can measure distances between given locations and return either the distance or the commute time. For this study's purpose, we decided to use solely the driving distance, as commute time can vary depending on different traffic conditions and other outside factors. Due to the large computational cost of OSRM, both datasets were required to be scaled down in size. We used random sampling, leaving 6293 considered residences for the Indiana dataset and 1649 house blocks in the California dataset. Moreover, to simulate weight for the California dataset, we define a weighting factor $w_i$ as

$$w_i = \frac{5}{I_i},$$

where $I_i$ is the annual income of the $i$th household given in units of $10,000. It can be readily seen that $w_i$ is equal to or greater than 1.25 and is larger for lower income households. Ideally, when calculating the weighted average distance between households and food pantries we would assign the weighting factor $w_i$

to the i*th* household, thereby prioritizing lower-income households. However, OSRM does not explicitly offer weighted averages, and we employ a trick of duplicating the i*th* household $w_i$ times (rounded to the closest integer) in the dataset, to equivalently obtain the weighted average eventually. As a result, we considered 3,398 total house blocks (1649 unique) in the California dataset after all modifications. After both datasets were prepared, the OSRM distance matrix function was run on both, yielding two distinct distance matrices that were used as inputs for the K-Medoids algorithm.

**2.2 K-Medoids Implementation**

According to Feeding America, food banks generally collect and sort food donations before distributing these resources to food pantries, who themselves are responsible for directly distributing food to their communities ("How Do Food Banks Work? | Feeding America," n.d.). To simulate this same logistical chain, we used K-Medoids in a two-level manner, where it is first run on the two overall datasets to determine optimal food bank locations, before being run in each individual cluster to determine optimal food pantry locations.

The algorithm for determining optimal food pantry locations is sketched in Fig. 2. K-Medoids begins by assigning K initial centroids (food pantry locations in this work) to a dataset (households in this work), with all centroids being existing data points. These initial centroids can be either randomly chosen or user picked. For this study's purpose, we chose the initial centroids as the first K items in the dataset. Subsequently, the algorithm has two main stages. The first stage is the assignment of all data points to their nearest centroid. The second stage attempts to determine whether a new centroid would be more optimal than an existing one, by randomly swapping a centroid to another data point in an existing cluster. If the new centroid is determined to be better at minimizing the total distance within the cluster, K-Medoids will keep the new centroid location and revert all other non-optimal changes. As some data points may become closer to a different cluster centroid than their last cluster after centroid movement, all data points will be reassigned to their new closest centroid. After that, K-Medoids will begin a new iteration of the second stage. This process of moving the centroids and reassigning the data points will continue until convergence is achieved, i.e., all data points stay in their same cluster, and all centroid positions are stable. All K-Medoids implementations were not given an iteration limit.

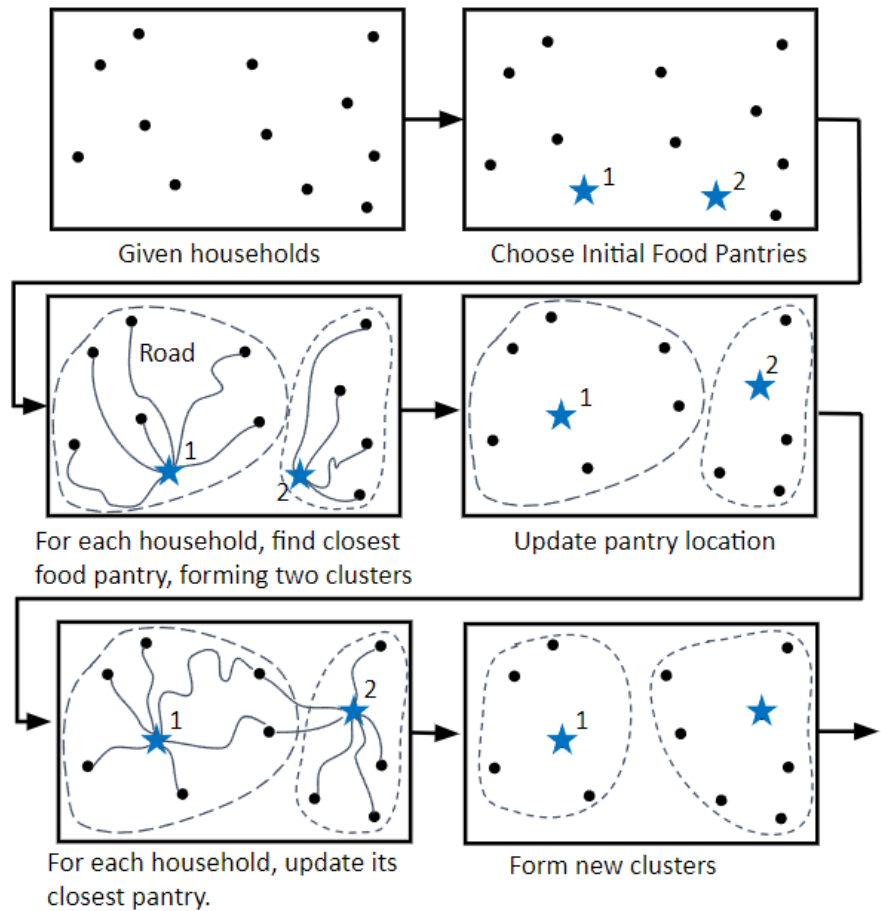

**Figure 2** An overview of the K-Medoids clustering method using a schematic example involving 12 houses and 2 food pantries.

Due to the nature of the weighted California dataset, lower income houses would be more prevalent in the dataset due to their duplicates. As a result, they were more heavily prioritized by the K-Medoids algorithm hence the food pantries were generally closer to them. The weighted dataset results were also compared to those unweighted and the effectiveness of weighing will be discussed in the results section.

For optimizing food pantry and bank locations in California, 57 pantries were generated to compare against 57 real food pantries, each partnered with one of the Feeding America affiliated food banks. 17 food bank locations were generated to compare against 17 real-food bank locations all affiliated with Feeding America. For the Indiana dataset, 176 total food pantries were generated to compare against 176 real affiliated pantries and 9 food bank locations were generated to compare against 9 Feeding America affiliated food banks. Due to the proximity of several K-Medoids food pantries and banks to their real counterparts, we gathered the real pantry data on a city-to-city basis. For California, these cities were Los Angeles, San Diego, San Jose, Fresno, and Oakland. For Indiana, they were Bloomington, Fort Wayne, Muncie, Merrillville, Lafayette, Terre Haute, Evansville, South Bend, and Indianapolis.

## 3 RESULTS

### 3.1 Food pantry location comparison

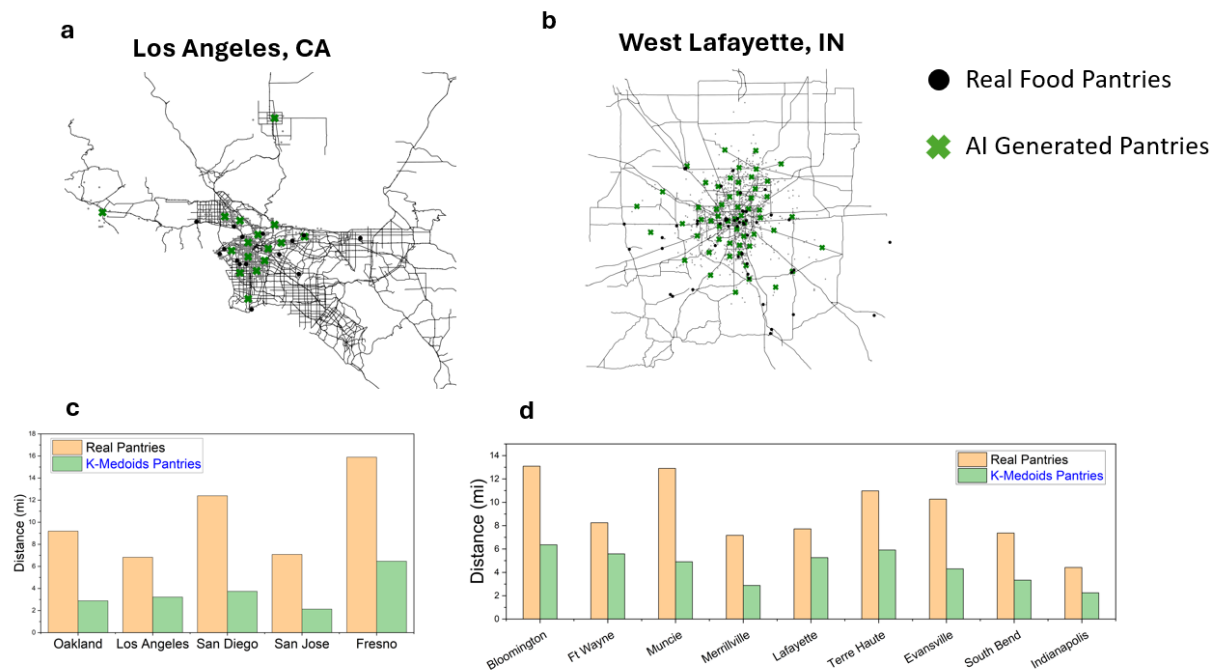

**Figure 3 (a)** Comparison of real and AI (K-Medoids) food pantry locations in Los Angeles, CA. **(b)** Comparison of real and AI food pantry locations in West Lafayette, IN. **(c)** Average household-pantry distance comparisons across cities in California. **(d)** Average distance comparisons across cities in Indiana.

Figures 3 (a) and (b) show examples of the final locations for K-Medoids food pantries compared to their real counterparts in Los Angeles, CA and West Lafayette, IN. In Los Angeles, the average distance between a house block and its closest real food pantry is 6.83 miles, compared to a K-Medoids average of 3.22 miles, yielding a significant saving of 3.61 miles or 52.9%. For West Lafayette, the real average distance is 4.41 miles compared to an average K-Medoids distance of 2.25 miles, yielding a significant saving of 2.16 miles or 49.0%. An interesting observation is that the K-Medoids food pantries look more scattered than their real counterparts, potentially offering better access to some households that were under-served by the real pantries. As seen from Figures 4 (c) and (d), city-wide average savings for each household ranged from 3.61 to 9.43 miles in California and 2.17 to 8 miles in Indiana. Typically, cities with less food pantries show larger average distance savings. Statewide, our method overall saved 19,432.22 total miles between households and food pantries in California and 22,181.42 miles in Indiana, with average savings at 5.72 and 3.52 miles, respectively.

### 3.3 Food bank location comparison

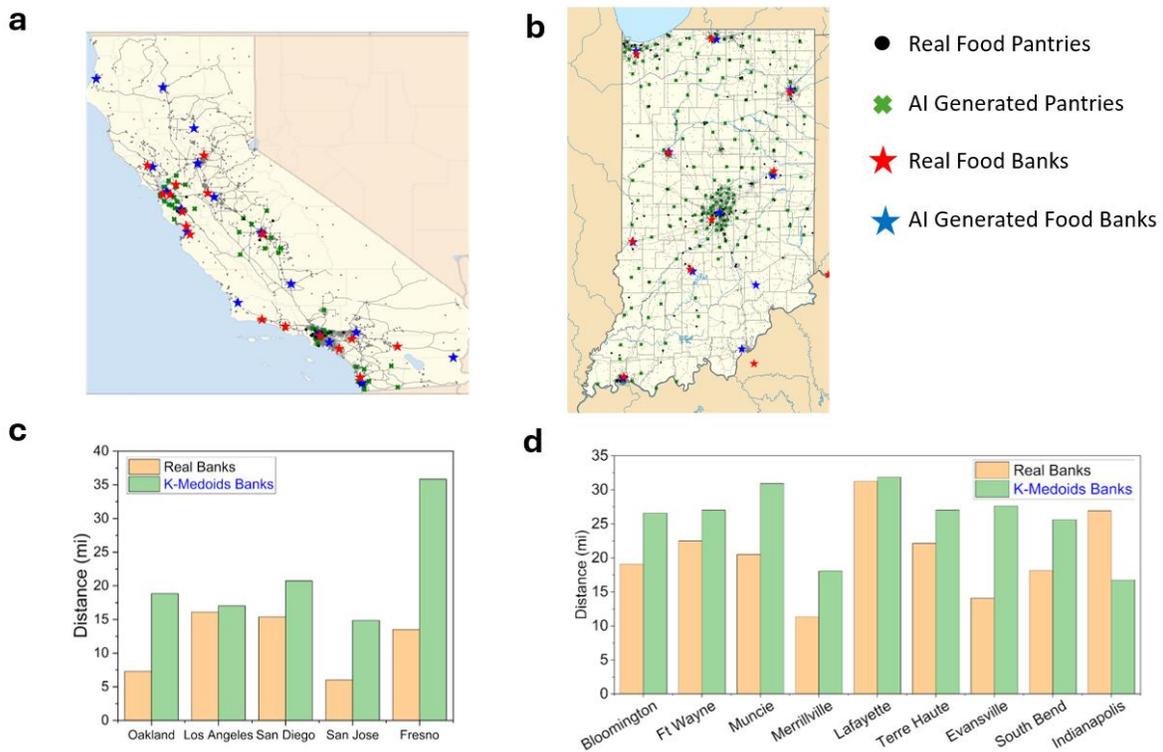

**Figure 4** (**a**) Optimized food bank locations of California dataset. (**b**) Optimized food bank locations of Indiana dataset. (**c**) Average food bank to pantry distance across cities in California. (**d**) Average food bank to pantry distance across cities in Indiana.

Figures 5 (a) and (b) show final K-Medoids food bank locations compared to their real counterparts. Interestingly, we found that on the food bank level, our method yielded a slight penalty in the distance between food pantries and their supplier food banks. Per pantry, this yields an average of 10.02 miles of penalty for California and 1.56 miles for Indiana. For California, there was 571.21 total miles of penalty, while for Indiana this penalty was 273.75 miles.

**4 DISCUSSION**

The purpose of this project was to develop a methodology that could factor in road distance and other factors to find optimal locations for food pantries and banks. Results show that our methodology, utilizing a two-level K-Medoids algorithm in conjunction with the Open Source Routing Machine, was effective in producing optimal food pantry and bank results that had superior overall locations when compared to real ones. As seen by Figure 3, our method saved significant total and average distance between households and their closest food pantries. The smaller average distance savings in Indiana was likely because of differences in state size.

While significant distance was saved on the second level, i.e., the household-pantry level, we found a slight penalty on the first pantry-bank level. However, since the number of households is much larger than the number of pantries, the total penalty distance on food banks to deliver to pantries was only a small fraction of the total saved distance for households to access pantries. As such, we believe that this methodology's benefits far outweigh any of the discovered side effects. Furthermore, the average distance penalty on food banks for both Indiana and California are less significant when considering that large food organizations such as Feeding America have both ample access to transportation (over 2400 available supply trucks) and funding (over $425 million in cash in 2023), especially when compared to food insecure households that may operate on limited transportation access and budget ("How does Feeding America respond to disasters? | Feeding America," 2023;"2023 Auditor's Report | Feeding America," 2023).

Future improvements could be made to this methodology by increasing its practicality. Food bank capacity was not considered in this project. However, food banks do not have infinite resources. Assigning maximum capacities to each food bank would further increase the practicality of such a program.

To summarize, we developed a machine learning method to optimize food bank and pantry locations. By using a two-level K-Medoids algorithm in conjunction with the Open Source Routing Machine, we were able to save significant road distance between households and their closest available food resources. Comparing to existing food pantry and bank locations, interestingly our program significantly prioritizes distance savings for households while bringing slight penalty for food banks. This work can not only help decrease America's rate of food insecurity by making food pantries more accessible, but also plan out food bank and pantry locations in countries where little to no food aid infrastructure currently exists.